\documentclass[10pt,twocolumn,letterpaper]{article}

\usepackage[pagenumbers]{cvpr} 

\usepackage{graphicx}
\usepackage{amsmath}
\usepackage{amssymb}
\usepackage{booktabs}

\usepackage[pagebackref,breaklinks,colorlinks]{hyperref}

\usepackage[capitalize]{cleveref}
\crefname{section}{Sec.}{Secs.}
\Crefname{section}{Section}{Sections}
\Crefname{table}{Table}{Tables}
\crefname{table}{Tab.}{Tabs.}

\def\cvprPaperID{*****} 
\def\confName{CVPR}
\def\confYear{2024}

\begin{document}

\title{3rd Place Solution for PVUW Challenge 2024: Video Panoptic Segmentation}

\author{Ruipu Wu$^{1,2}${\qquad}Jifei Che$^{1}${\qquad}Han Li$^{2}${\qquad}Chengjing Wu$^{1}${\qquad}Ting Liu$^{1}${\qquad}
Luoqi Liu$^{1}$\vspace{3mm}\\
$^{1}$MT Lab, Meitu Inc\qquad$^{2}$Beihang University 
}
\maketitle

\begin{abstract}
Video panoptic segmentation is an advanced task that extends panoptic segmentation by applying its concept to video sequences. In the hope of addressing the challenge of video panoptic segmentation in diverse conditions, We utilize DVIS++ as our baseline model and enhance it by introducing a comprehensive approach centered on the query-wise ensemble, supplemented by additional techniques. Our proposed approach achieved a VPQ score of 57.01 on the VIPSeg test set, and ranked 3rd in the VPS track of the 3rd Pixel-level Video Understanding in the Wild Challenge.
\end{abstract}

\section{Introduction}
\label{sec:intro}
In order to unify the video tasks including recognition, detection, tracking and segmentation, Kim \textit{et al.} propose Video Panoptic Segmentation(VPS) by extending the concept of panoptic segmentation from image domain to video domain~\cite{vps}. 
Video panoptic segmentation can be seen as a fusion of video semantic segmentation and video instance segmentation \cite{vis}. More specifically, in a VPS task, each pixel in an image is required to be assigned both a semantic label and an instance ID. 
The VPS task has received a lot of attention due to its wide application in downstream tesks including video editing, video understanding and autonomous driving.

Early studies in VPS domain focused on adapting image-based panoramic segmentation methods to the video domain by leveraging temporal consistency across video frames. VPSNet\cite{vps} integrated a temporal pixel-level fusion module and an object tracking head into a image panoptic segmentation network UPSNet\cite{upsnet} to achieve panoptic video results. Video K-Net\cite{video-knet} utilizes cross-temporal kernel interaction based on K-Net \cite{knet} to simultaneously segment and track both "things" and "stuff" in videos. ViP-DeepLab\cite{vip-deeplab}, adapted from Panoptic-DeepLab\cite{panoptic-deeplab} achieves temporal alignment by utilizing offset predictions for consecutive frames, ensuring consistent instance grouping across the video sequence. Considering that earlier methods necessitated the use of several distinct networks and sophisticated post-processing, Max-DeepLab\cite{max-deeplab} proposed the first end-to-end model for panoptic segmentation that applies a pixel and memory dual-path modeling. MaXTron \cite{maxtron} integrates a mask transformer with trajectory attention to conduct VPS, enhancing temporal coherence with its within-clip and cross-clip tracking modules. 

\begin{figure}[t]
\centering
\includegraphics[width=1\columnwidth]{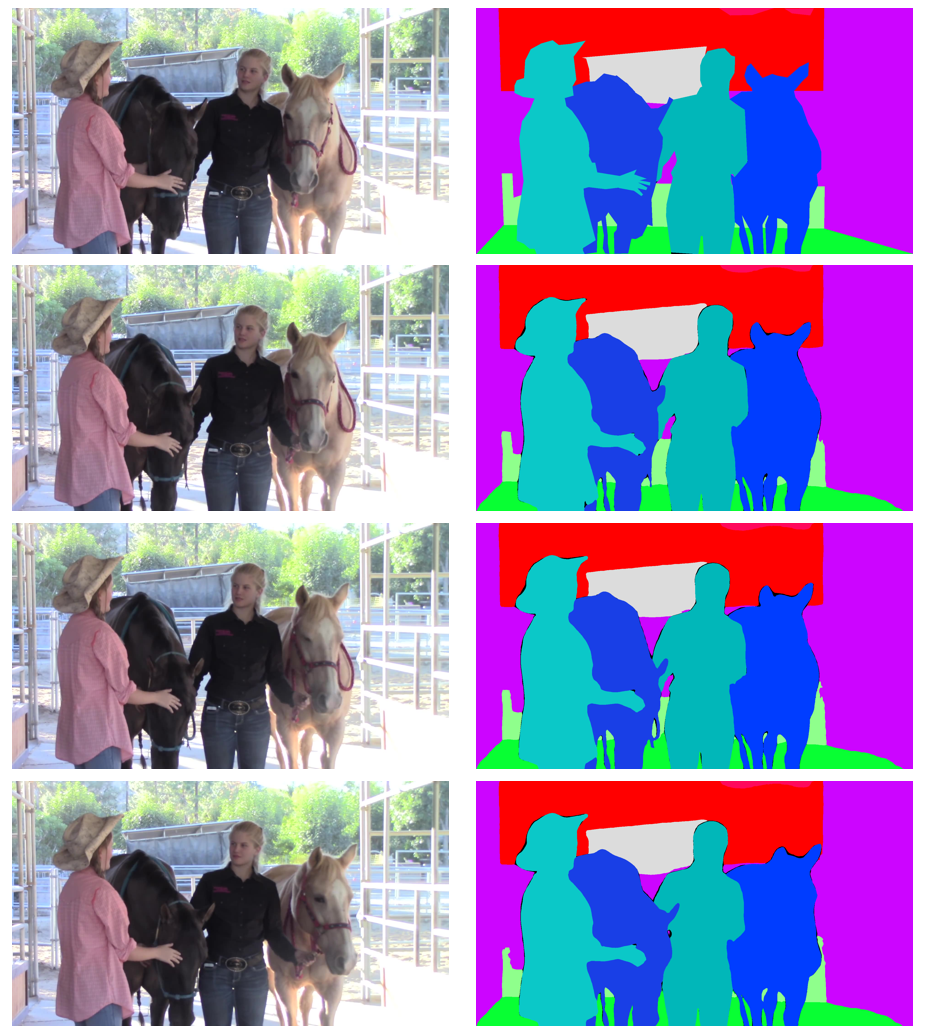}
\vspace{-0.5cm}
\caption{Examples of the VIPSeg dataset~\cite{vipseg}.}
\label{fig:dataset}
\end{figure}

Recently, DVIS \cite{dvis} introduces a decoupled framework that divides video segmentation into three independent sub-tasks: image segmentation, referring tracking and temporal refinement. Due to its outstanding performances, DVIS has achieved first place in the VPS track of 2nd PVUW Challenge in CVPR 2023. DVIS++\cite{dvis++} introduced denoising training and contrastive learning strategies into DVIS, resulting in a significant improvement and achieving state-of-the-art performance on VIPSeg\cite{vipseg} dataset. 

In this paper, we utilized DVIS++ as our baseline, and tried several techniques to deal with the inconsistency in the predicted results. Note that DVIS++ follows the set prediction paradigm, which generates superfluous queries, each representing a binary segmentation mask for a single class, usual practices of test time augmentation and model ensemble cannot be applied here directly. Therefore, we proposes an query-wise ensemble method to solve this problem.    

With the help of the query-wise ensemble method mentioned above and other additional techniques, our approach achieved 57.01 VPQ on the test set of VIPSeg dataset and ranked 3rd in the VPS track of the 3nd PVUW Challenge in CVPR 2024.

\section{Baseline Model}
\begin{figure*}[ht]
    \centering
    \includegraphics[width= 16cm]{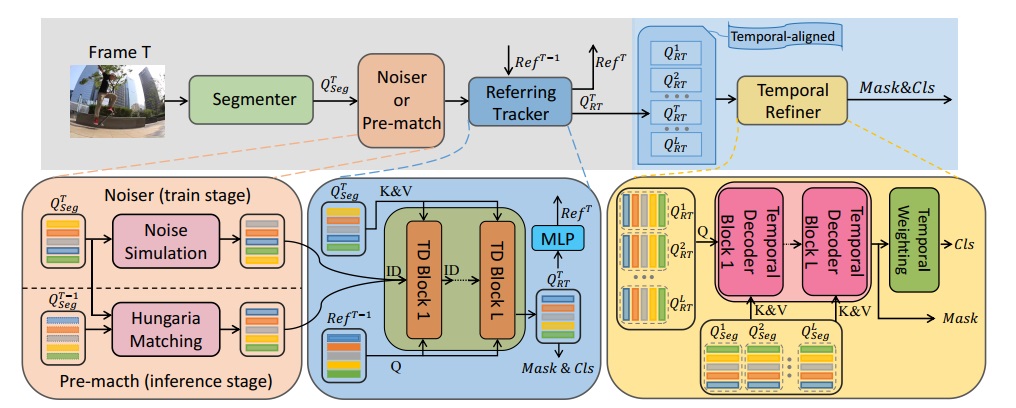}
    \caption{Architecture of DVIS++\cite{dvis++}.}
    \label{fig:dvis_plus_framework}
\end{figure*}

In this section, we will introduce our baseline model DVIS++.
DVIS++ has 3 decouple modules: segmenter, referring tracker and temporal refiner as shown in \cref{fig:dvis_plus_framework}. In this section, we will present these modules sequentially.

\subsection{Segmenter}
Segmenter serves as a image panoptic segmentation module. In our case,
DVIS++ employ Mask2Former\cite{Mask2Former} as the segmenter. Mask2Former is a versatile image segmentation architecture that outperforms specialized architectures across various segmentation tasks, all while ensuring straightforward training for each specific task. It is built on a simple meta-architecture consisting of a backbone, a pixel decoder, and a transformer decoder.

\subsection{Referring Tracker}
The referring tracker utilizes the modeling paradigm of referring denoising to address the inter-frame correlation task. Its primary goal is to utilize denoising operations to refine initial values and produce more precise tracking outcomes.

The referring tracker comprises a sequence of $L$ transformer denoising (TD) blocks, each composed of a referring cross-attention (RCA), a standard self-attention, and a feed-forward network (FFN). It takes object queries $ \left\{ Q^i_{seg}| i \in [1,T] \right\}$ generated by the segmenter as input and produces object queries $ \left\{ Q^i_{Tr}| i \in [1,T] \right\}$ for the current frame that corresponds to objects in the previous frame. In this context, T represents the length of the video.

Firstly, the Hungarian matching algorithm \cite{hungarian} is used to match the $Q_{seg}$ of adjacent frames, as done in \cite{MinVIS}, having $\widetilde{Q}_{seg}$ as results. Here, $\widetilde{Q}_{seg}$ represents the matched object query generated by the segmenter. $\widetilde{Q}_{seg}$ serves as the initial query for the reference tracker, albeit with noise. To remove noise from the initial query $\widetilde{Q}_{seg}^i$ of the current frame, the reference tracker utilizes the denoised object query $Q_{Tr}^{i-1}$ from the previous frame as a reference.

Then, $\widetilde{Q}_{seg}^i$ is processed in the TD block, where the denoising process is performed using RCA, resulting in the output $Q^i_{Tr}$. RCA effectively leverages the similarity between object representations of adjacent frames while mitigating potential confusion caused by their similarity. To address the issue of ambiguous object representation initialization from the previous frame \cite{GenVIS}, RCA incorporates an identity (ID) mechanism, effectively exploiting the similarity between the query (Q) and key (K) to generate accurate outputs. 

Finally, the denoised object query $Q^i_{Tr}$ is used as input for both the class head and mask head. The class head generates the category output, while the mask head produces the mask coefficient output.

\subsection{Temporal Refiner}
Previous offline video segmentation methods have been limited by the inadequate utilization of temporal information in tightly coupled networks, while current online methods lack a refinement step. To address these challenges, we propose an independent temporal refiner module. This module efficiently utilizes temporal information across the entire video to refine the output generated by the referring tracker.

The architecture of the temporal refiner plays a crucial role in enhancing the model's utilization of temporal information. It takes the object query $Q_{Tr}$ from the reference tracker as input and produces the refined object query $Q_{Rf}$ by aggregating temporal information from the entire video. The temporal refiner consists of L temporal decoder blocks connected in a cascaded operation. Each block comprises a short-term temporal convolutional block and a long-term temporal attention block, leveraging motion information and integrating information from the entire video, respectively, through 1D convolutions and standard self-attention.

Finally, the mask head generates mask coefficients for each object in every frame using the refined object query $Q_{Rf}$. Additionally, the class head predicts the class and score of each object across the entire video using the temporal weights of $Q_{Rf}$.

\section{Our Solution}
In this section, we will introduce the improvements we have made compare to the baseline DVIS++.

\subsection{Query-wise ensemble}
In order to adapt the test time augmentation and the model ensemble methods to the set prediction paradigm in DVIS++, we proposed the query-wise ensemble. More concretely, given the original queries and the queries generated from the augmented videos or by the supplementary models, we compare the binary mask corresponding to each supplementary query with the original ones, if the IOU between the current binary mask and one of the original binary mask is bigger than 0.5, then we average the mask logits and class logits of the two queries. Otherwise, the supplementary query is added to the original query set.

In the above methods, if the supplementary query matches with one of the original query, we will use it to refine the corresponding original query. Otherwise, we can reckon that the query correspond to a missed object, and thus we can leverage this query to supplement the original result. Ideally, our method will refine the edge of the existing masks and assign new masks to the pixel of void class.

\subsection{Other Operations}
We also tried to use DEVA\cite{deva}, a post-process temporal alignment module, to refine the segmentation results. However, the results were not up to our expectations, leading us to exclude it from our final inference process.

\section{Experiment}
\subsection{Dataset}
\textbf{VIPSeg.}\ VIPSeg \cite{vipseg}provides 3,536 videos and 84,750 frames with pixel-level panoptic annotations, covering a wide
range of real-world scenarios and categories, which is the first attempt to address the challenging task of video panoptic segmentation in the wild by considering a variety of scenarios.VIPSeg is divided into train set, validation set and test set each contains 2, 806/343/387 videos respectively. VIPSeg showcases a variety of real-world scenes across 124 categories, consisting of 58 categories of 'thing' and 66 categories of 'stuff'. Due to limitations in computing resources, all the frames in VIPSeg are resized into 720P (the size of the short side is resized to 720) for training and testing. 

\subsection{Implementation Details}
In our approach, we retrain the model using the merged training and validation sets of the VIPSeg dataset, following the default setting provided by the author as presented below. We divide it into three stages to train the segmenter, referring tracker, and temporal refiner.
In the first stage, we utilize the pretrained offline model weight which is fine-tuned from the COCO\cite{coco} pretrained segmenter weight and employs VIT-L\cite{ViT-adapter} as the backbone segmenter. In the following stages, we freezed the module trained before, and retrained the corresponding module. Besides, Training is carried out for 20k iterations with a batch size of 8 on 8 NVIDIA Tesla V100, and the learning rate is decayed by 0.1 at 14k iterations.

As for the test time augmentation, we tried horizontal flip, brightness augmentation, contrast augmentation and multi-scale ensemble which combines the result of the 720p and 800p videos. We also attempted to ensemble the results of MaxTron \cite{maxtron} and UniVis \cite{li2024univs}, but encountered difficulties due to the corrupted weights of MaxTron. 

\subsection{Comparison with Other Methods}
In the 3rd PVUW Challenge, we ranked third in both the test phase and development. Our method achieved a VPQ of 54.55 in the development phase and 57.01 in the test phase, as shown in the \cref{tab:development} and \cref{tab:test}. The comparison between our method and baseline can be found in \cref{tab:comparison}.

\begin{table}
  \centering
  \resizebox{1.0\columnwidth}{!}{
  \begin{tabular}{ccccccc}
    \toprule
    Team & VPQ & VPQ1 & VPQ2 & VPQ4 & VPQ6 & STQ \\
    \midrule
    SiegeLion&	56.3598& 57.1408& 56.4636& 56.0302& 55.8046& 0.5252   \\
     kevin1234&	55.6940 & 56.4139& 55.8574& 55.3925& 55.1122& 0.5190   \\
    \color{blue}{Reynard}&	\color{blue}{54.5464}  & \color{blue}{55.2727} & \color{blue}{54.6924} & \color{blue}{54.2534} & \color{blue}{53.9672} & \color{blue}{0.5166}    \\
    ipadvideo&	54.2571 & 54.9604& 54.4390& 53.9786 & 53.6504 & 0.5093   \\
    zhangtao-whu&	52.7673 & 53.3162& 52.9243& 52.5669 & 52.2618 & 0.5016   \\
    \bottomrule
  \end{tabular}
  }
  \caption{Ranking results of leaderboard during the development phase.}
  \label{tab:development}
\end{table}

\begin{table}
  \centering
  \resizebox{1.0\columnwidth}{!}{
  \begin{tabular}{ccccccc}
    \toprule
    Team & VPQ & VPQ1 & VPQ2 & VPQ4 & VPQ6 & STQ \\
    \midrule
    kevin1234&	58.2585& 59.1009& 58.5042& 57.9007& 57.5283& 0.5434   \\ 
     SiegeLion&	57.1188 & 58.2143& 57.4119& 56.6798& 56.1691& 0.5397   \\
    \color{blue}{Reynard} & \color{blue}{57.0114} & \color{blue}{57.8900} & \color{blue}{57.2240} & \color{blue}{56.6509} & \color{blue}{56.2807} & \color{blue}{0.5343}   \\
    ipadvideo&	28.3810 & 29.1165& 28.6770& 28.1028 & 27.6277 & 0.2630   \\
    JMCarrot&	22.1060 & 23.8061& 22.8334& 21.4910 & 20.2935 & 0.2603   \\
    \bottomrule
  \end{tabular}
  }
  \caption{Ranking result of leaderboard during the test phase.}
  \label{tab:test}
\end{table}

\begin{table}
  \centering
  \resizebox{1.0\columnwidth}{!}{
  \begin{tabular}{ccccccc}
    \toprule
    Method & VPQ & VPQ1 & VPQ2 & VPQ4 & VPQ6 & STQ \\
    \midrule
    Baseline&	56.1997 & 57.1684 & 56.4638& 55.8131& 55.3535& 0.5293  \\
    Final results&	57.0114 & 57.8900 & 57.2240 & 56.6509 & 56.2807 & 0.5343   \\
    \bottomrule
  \end{tabular}
  }
  \caption{Comparison between our method and baseline on test set.}
  \label{tab:comparison}
\end{table}

\section{Conclusion}

In this paper, we introduced query-wise ensemble to DVIS++ to improve the temporal consistency of the results and alleviate the issue of missed segmentation. As a result, we get the 3nd place in the VPS track of the PVUW Challenge 2024, scoring 54.55 VPQ and 57.01 VPQ in the development and test phases respectively.
{\small
\bibliographystyle{ieee_fullname}
\bibliography{reference}
}

\end{document}